\documentclass[environsciproc,article,submit,pdftex,moreauthors]{Definitions/mdpi}

\preto{\abstractkeywords}{\nolinenumbers}

\usepackage{graphicx}
\usepackage{pgfplots}
\usepackage{graphicx}
\usepackage{tabularx}
\usepackage{mathtools}
\usepackage{tabulary}
\usepackage{comment}
\firstpage{1}
\pubvolume{1}
\issuenum{1}
\articlenumber{0}
\pubyear{2023}
\copyrightyear{2023}
\datereceived{ }
\daterevised{ } 
\dateaccepted{ }
\datepublished{ }
\usepackage{pifont}
\newcommand{\cmark}{\ding{51}}%
\newcommand{\xmark}{\ding{56}}%
\hreflink{https://doi.org/} 

\Title{Empirical Study of PEFT techniques for Crop Type Segmentation}

\TitleCitation{Empirical Study of PEFT techniques for Winter Wheat Segmentation}

\Author{Mohamad Hasan Zahweh$^{1,2}$, Hasan Nasrallah$^{2}$, Mustafa Shukor $^{3}$, Ghaleb Faour$^{2}$ and Ali J. Ghandour\orcidA{} $^{2}$*}

\AuthorNames{Mohamad Hasan Zahweh, Hasan Nasrallah, Mustafa Shukor, Ghaleb Faour and Ali J. Ghandour}

\AuthorCitation{Zahweh, M.; Nasrallah, H.; Shukor, M.; Faour, G.; Ghandour, A.}

\address{%
$^{1}$ \quad Lebanese University;\\
$^{2}$ \quad National Center for Remote Sensing - CNRS, Lebanon;\\
$^{3}$ \quad Sorbone University;}

\corres{Corresponding author: Ali J. Ghandour, aghandour@cnrs.edu.lb;}

\abstract{Parameter Efficient Fine Tuning (PEFT) techniques have recently experienced significant growth and have been extensively employed to adapt large vision and language models to various domains, enabling satisfactory model performance with minimal computational needs. Despite these advances, more research has yet to delve into potential PEFT applications in real-life scenarios, particularly in the critical domains of remote sensing and crop monitoring.
In the realm of crop monitoring, a key challenge persists in addressing the intricacies of cross-regional and cross-year crop type recognition. The diversity of climates across different regions and the need for comprehensive large-scale datasets have posed significant obstacles to accurately identify crop types across varying geographic locations and changing growing seasons. This study seeks to bridge this gap by comprehensively exploring the feasibility of cross-area and cross-year out-of-distribution generalization using the State-of-the-Art (SOTA) wheat crop monitoring model.
The aim of this work is to explore efficient fine-tuning approaches for crop monitoring. Specifically, we focus on adapting the SOTA TSViT model, recently proposed in CVPR 2023, to address winter wheat field segmentation, a critical task for crop monitoring and food security, especially following the Ukrainian conflict, given the economic importance of wheat as a staple and cash crop in various regions. This adaptation process involves integrating different PEFT techniques, including BigFit, LoRA, Adaptformer, and prompt tuning, each designed to streamline the fine-tuning process and ensure efficient parameter utilization.
Using PEFT techniques, we achieved notable results comparable to those achieved using full fine-tuning methods while training only a mere 0.7\% parameters of the entire TSViT architecture. More importantly, we achieved the claimed performance using a limited subset of remotely labeled data. The in-house labeled data-set, referred to as the Lebanese Wheat dataset, comprises high-quality annotated polygons for wheat and non-wheat classes for the study area in Beqaa, Lebanon, with a total surface of 170 km², over five consecutive years from 2016 to 2020. Using a time series of multispectral Sentinel-2 images, our model achieved a 84\% F1-score when evaluated on the test set, shedding light on the ability of PEFT to drive accurate and efficient crop monitoring, designed mainly for developing countries characterized by limited data availability.
Our code is publicly available at this \href{https://github.com/geoaigroup/GEOAI-ECRS2023}{Repo}.
}


\begin{document}

\section{Introduction}

\begin{figure}[t]
\begin{center}
\includegraphics[scale=0.5]{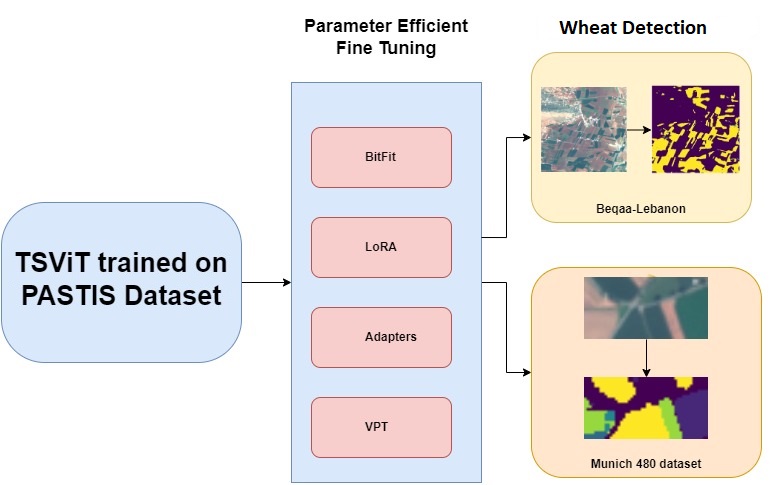}
\end{center}
\caption{Large transfomer-based model PEFT-ing for down-stream task: TSViT model is pre-trained on PASTIS dataset and then multiple tuning techniques are applied for winter wheat crop detection using Beqaa-Lebanon and Munich 480 datasets.}
\label{fig:architecture}
\end{figure}

Foundation models, mostly based on transformer architecture, have demonstrated impressive success in many domains and benchmarks. However, the exponential increase in the number of parameters makes training these models extremely expensive and challenging. This motivated researchers to propose more efficient approaches to fine-tune and adapt these models to downstream tasks.

Parameter-Efficient FineTuning (PEFT) encompasses a family of approaches that keep most of the parameters of pre-trained models frozen and train only part or additional parameters. PEFT approaches have been extensively investigated and used in the NLP community and recently in the vision community. However, most of the work still considers general academic benchmarks and classical tasks.

In this work, we explore efficient fine-tuning approaches for crop monitoring. 
Specifically, we consider the recent state-of-the-art (SoTA) TSViT model that relies on time series of satellite images. Rather than fully tuning all model parameters, we fine-tune only a small number, using different techniques such as bias tuning, Adapters and LOw-RAnk adaptation as shown in Figure~\ref{fig:architecture}. This paper contribution is two-folds:
\begin{itemize}
    \item We explore different PEFT techniques to efficiently adapt pre-trained models for crop type segmentation, and identify the most important parameters for each PEFT method to achieve the best performance.
    \item We experiment with different datasets from different countries like Germany and Lebanon and consider a realistic and challenging setup where we test the model on different year and regions.

\end{itemize}

\section{Framework}

In this work, we focus on efficient fine tuning of large transformer-based model for winter wheat crop segmentation, as shown in Figure~\ref{fig:architecture}. The Temporal Spatial Vision Transformer (TSViT)~\cite{tarasiou2023vits} is a recent state-of-the-art ision transformer that achieved SOTA for satellite image time series in the PASTIS dataset~\cite{PASTIS_dataset}. The model consists of a temporal transformer followed by a spatial transformer. The architecture is based on common practices used for video processing, where the model used temporal encoding before spatial encoding. Given its state-of-the-art performance in crop field segmentation, we choose TSViT as the main model upon which we conduct our experiments and comparisons.

\subsection{PEFT techniques}
\label{section_peft}
To avoid training all model parameters, especially large transformer models, great effort has been put into promoting efficient fine-tuning approaches~\cite{lialin2023scaling}. Most of this work is focused on Large-Language Models (LLMs)~\cite{lester2021powerprompttuning,houlsby_PEFT,hu2021lora,prefix_tuning,karimi2021compacter}. Some effort has been made to address vision model efficient tuning ~\cite{jia2022vpt,chen2022adaptformer}, however, very little work has considered PEFT techniques for remote sensing satellite images applications~\cite{yuan2023parameter}.

Parameter Efficient Fine Tuning (PEFT) techniques focus on training the least size of parameters in the most efficient way possible. These techniques achieve results equivalent to or better than full fine-tuning, reducing the cost of both training and storing the model with minimal loss (and sometimes non).

In our work, we investigate the effectiveness of the following PEFT techniques for the task of crop field segmentation given a time series of Sentinel-2 images: \textit{(i)} BitFit, \textit{(ii)} Visual Prompt Tuning, \textit{(iii)} LoRA, \textit{(iv)} Adapter.

\textbf{BitFit}~\cite{zaken2021bitfit} is a technique that focuses on training the Bias only, rather than the whole model. BitFit requires the least number of trainable parameters among other PEFT methods (note that head tuning requires fewer trainable parameters than BitFit).

\textbf{Visual Prompt Tuning (VPT)}~\cite{jia2022visual} is a PEFT technique that adds additional prompt parameters concatenated to the input of the transformer that can help the model understand the different tasks and achieve better results. This prompt can be added to the first transformer encoder layer only, which is called \textbf{Shallow} prompt tuning, or added to each transformer encoder layer, which is known as \textbf{Deep} prompt Tuning. Furthermore, given that the TSViT architecture is uniquely characterized by the presence of two transformers, one focused on temporal information followed by another on spatial data, it opened up an opportunity to experiment with dual prompts.

\textbf{LoRA}~\cite{hu2021lora} aims to train a smaller coinciding model with a low intrinsic rank $r$. Low-Rank Adaption (LoRA) freezes the original weight matrix \(W_0\)  and trains a separate update matrix that is formed by the product of two separate matrices $A$ and $B$. In our LoRA experiments, we investigate the effect of the following hyperparameters of the additional modules: \textit{(i)} $rt$: rank of the temporal transformer LoRA layers, \textit{(ii)} $rs$: rank of the spatial transformer LoRA layers, and \textit{(iii)} $rr$: rank of the rest of Lora layers.

\textbf{Adapter}~\cite{NEURIPS2022_69e2f49a} tuning is a method that aims to achieve efficient model adaptation by introducing two additional layers in the feed-forward steps. It can be added in series or in parallel with the feed forward layer.

Given the dual nature of the TSViT, our attention was primarily divided between two critical hyperparameters: the spatial adapter dimension and the temporal spatial dimension. The former pertains to the scale and complexity of spatial data processing, while the latter concerns the interpretation of temporal data. These dimensions essentially determine the granularity and depth of information processing within the transformer.

\subsection{Baselines}
Before the analytical comparison of our results, it is necessary to establish clear baselines. These baselines will serve as foundational benchmarks against PEFT. To ensure a logical comparison, we have incorporated a blend of traditional and specialized fine-tuning techniques. The techniques selected for our baseline are as follows:
\begin{itemize}
\item \textbf{Training from Scratch}: In this approach, the TSViT model was trained without using any prior weights.
 The goal of this strategy was to discern the innate potential of the TSViT architecture without the influence of fine-tuning.
\item \textbf{Full Fine-Tuning}: Given a pre-trained model, we apply transfer learning by training all model parameters on a new data set using a low learning rate. It represents an aspirational benchmark: any PEFT technique that can perform as well or better than full fine-tuning would be deemed successful.

\item \textbf{Head Fine-Tuning}: This technique introduces a layer to the front of the model, which then undergoes training. It is minimalist, targeting only the initial aspects of the model and setting the minimal performance expectation. Any PEFT technique that underperforms compared to this baseline would need re-evaluation.

\item \textbf{Token Tuning}: TSViT stands out due to its reliance on temporal tokens for segmentation. This intrinsic characteristic opens up the possibility of a unique fine-tuning technique: Token Tuning. By manipulating the temporal tokens, one can effectively alter the output classes. It is a nuanced method, specifically for the TSViT model.

\end{itemize}

\subsection{Datasets}

\label{sec:dataset}
We used two datasets in this study: \textit{(i)} Beqaa-Lebanon that we prepared in-house for the scope of this work and \textit{(ii)} Munich 480 dataset.

\textbf{Beqaa-Lebanon} data set was labeled for this project, where we annotated a weakly supervised data set of approximately 2,000 wheat parcels per year from 2016 to 2019. We then labeled an equivalent area of negative samples (non-wheat areas) per year to balance our data set. The total annotated area per year (from 2016 to 2019) is 170 $km^2$. Finally, we thouroughly annotated the year 2020 as a test set of surface area equal to 981 $km^2$, with about 16\% positive samples (wheat). In all our experiments, we used the annotated area of years 2016 to 2019 as our training set. We also split year 2020 into four tiles, where one of them was used for validation purpose, and the remaining three as a test set. All experimental results are based on the test set.

The study area is divided into $24*24$ tiles as in the paper~\cite{tarasiou2023vits}, which means that each input contains \textit{T*24*24*C}, where $T$ is the number of images that make up a time series of Sentinel-2 images for each wheat season, and $C$ is the number of Sentinel-2 bands used. We set $T = 9$, corresponding to a series of images from November to July, and $C = 10$.

\textbf{Munich 480} dataset~\cite{russwurm2018multi} covers an area of 102 $km$ * 42 $km$ in Munich, Germany. It consists of 30 images per year split into $48*48$ tiles made of all 13 Sentinel-2 bands and covering 27 different crop classes. We splitted images into $24*24$ tiles and used 60\%-20\%-20\% ratio for train-validation-test sets distribution.

\section{Experimental Results}
In this section, we will discuss the details of the implementation of the model, as well as the process of training or tuning the model using different techniques and baselines. In all our experiments, we used Adam optimizer, trained each model for 20 epochs, set batch size to 16 and kept a constant learning rate throughout the entire training process. All experiments were implemented using the Pytorch library on a single Nvidia GeForce TiTan XP 12GB GPU. 

\begin{figure}[t]
\begin{center}
\includegraphics[scale=0.15]{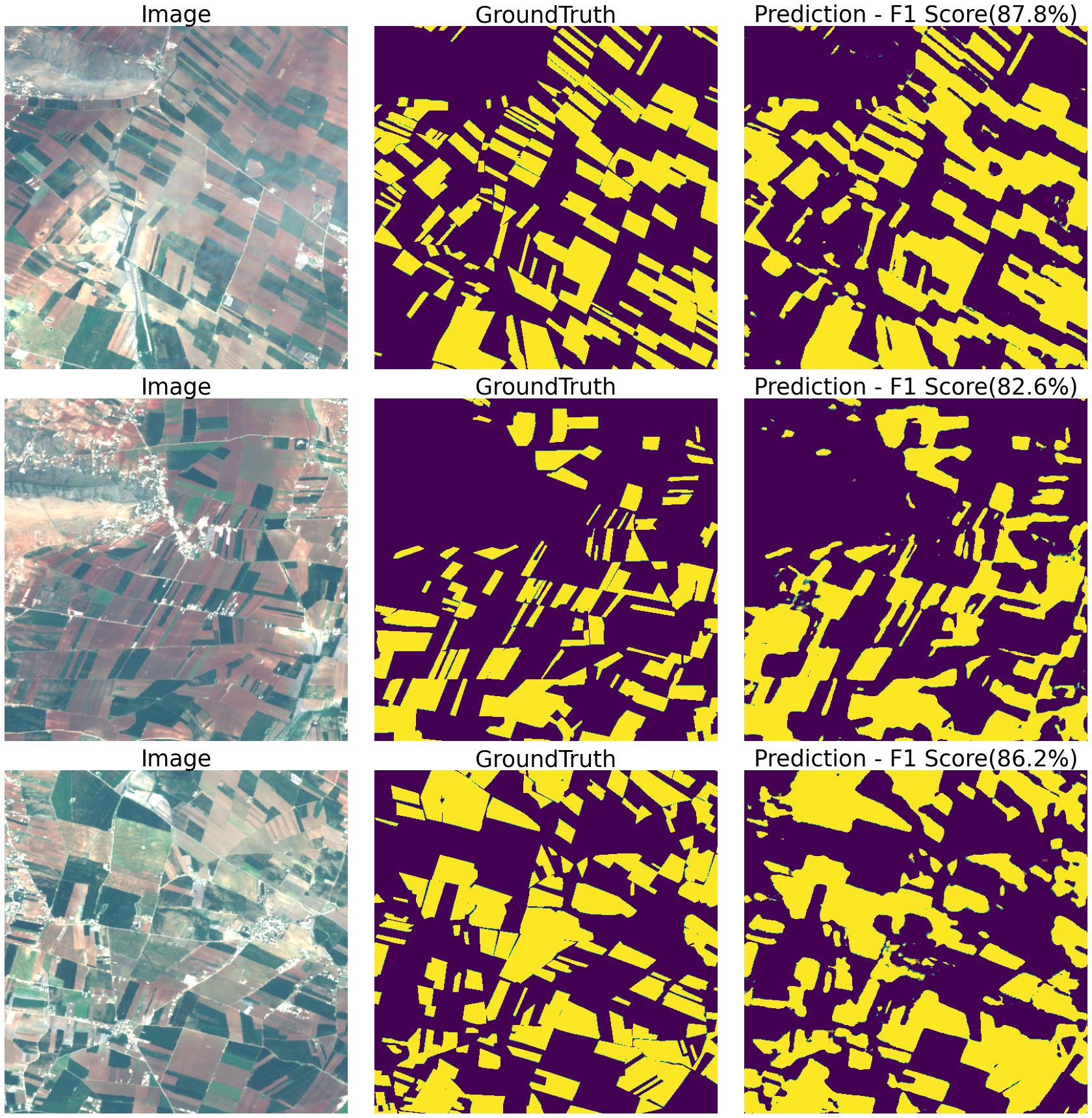}
\end{center}
\caption{Best model predictions and F1-scores for some tiles, in addition to the ground truth.}
\label{viz_preds}
\end{figure}

In the following subsections, we discuss and compare the results of different PEFT techniques using the Beqaa-Lebanon dataset. For comparison, F1-score is used as our main metric. F1-score measures a model’s accuracy by combining precision and recall scores. Cross-validation was performed after each epoch, and the model with the best validation score was saved. As shown in Figure~\ref{viz_preds}, the predictions of our best model achieve high F1-scores, in the order of ~85\%.

\begin{table}[h]
\centering
\begin{tabular}{l|cc}
\toprule
Method &
Trainable Parameters (\%) &
F1-score \\ \midrule
BitFit-Partial Bias & \bf{0.29} & 83.0 \\
BitFit-Full Bias & 0.54 & 83.9\\
VPT   	  & \bf{0.29}&  83.5 \\
LoRA  	  & 5.87  & 84.76\\
AdaptFormer     & 1.09  & \bf{85.0} \\ \midrule
Head tune   & 0.05  & 56.0 \\
Full finetuning  & 100   & 84.3\\
\bottomrule
\end{tabular}
\caption{Performance of various PEFT techniques and the baselines in term of F1-score and percentage of trainable parameters on Beqaa-Lebanon dataset. Adaptformer offers the best performance (even better than full fine-tuning) with an F1-score of 85\%.}
\label{table:general_comparison}
\end{table}

\subsection{BitFit and LoRA}
\textbf{BitFit} is the simplest PEFT technique but showed promising results. As shown in Table \ref{table:general_comparison}, partial- and full-bias PEFT-ing of the TSVit model on the Beqaa-Lebanon dataset can achieve an F1 score of \textbf{ 83\% and 83. 9\%}, respectively. These results are comparable to full fine-tuning while only training \textbf{0.29\% and 0.54\%} of the model parameters, respectively.

\begin{table}[]
\begin{tabular}{lccccc}
\hline
 & \multicolumn{1}{c}{Temporal} & \multicolumn{1}{c}{Spatial} & \multicolumn{1}{c}{External} & \multicolumn{1}{c}{Deep} & \multicolumn{1}{c}{F1-Score(\%)} \\
  & \multicolumn{1}{c}{Dimension} & \multicolumn{1}{c}{Dimension} & \multicolumn{1}{c}{Prompt} & \multicolumn{1}{c}{Prompt} & 
 \\ \hline
Series 1 & 4 & 4 & \xmark & \cmark & 82.12\\
         & 8 & 8 & \xmark & \cmark & 82.88 \\
         & 16 & 16 & \xmark & \cmark & 82.35 \\
         \hline
Series 2 & 4 & 4 & \cmark & \cmark & 82.16 	\\
         & 8 & 8 & \cmark & \cmark & 82.75 	\\
         & 16 & 16 & \cmark & \cmark & \textbf{83.50}  	\\\hline
Series 3 & 8 & 8 & \cmark & \xmark & 81.36 	\\
         & 4 & 4 & \cmark & \xmark & 79.00 	\\
         \hline
Series 4 & 8 & 0 & \cmark & \cmark & \textbf{83.50} \\
         & 0 & 8 & \cmark & \cmark & 69.73 \\
         \hline
\end{tabular}
\caption{Visual Prompt Tuning Experimentation Table}
\label{table:vpt}
\end{table}

For \textbf{LoRA}, we conducted four series of experiments to assess the influence of hyperparameters discussed in Section~\ref{section_peft}. The best results are obtained when all intrinsic ranks are equal. This observation is best explained by the fact that the original and downstream tasks are the same here (crop-field segmentation), and thus the LoRA training should be balanced to match the flow of the model. The best results of LoRA tuning can be reached by balancing all modules' ranks where the highest F1-score of 84.9\% is achieved with $rt=rs=rr=4$. Detailed results are not included here for space limitations.

\subsection{VPT and Adaptformer}

\begin{figure}[]
\begin{center}
    \includegraphics[width=0.9\linewidth]{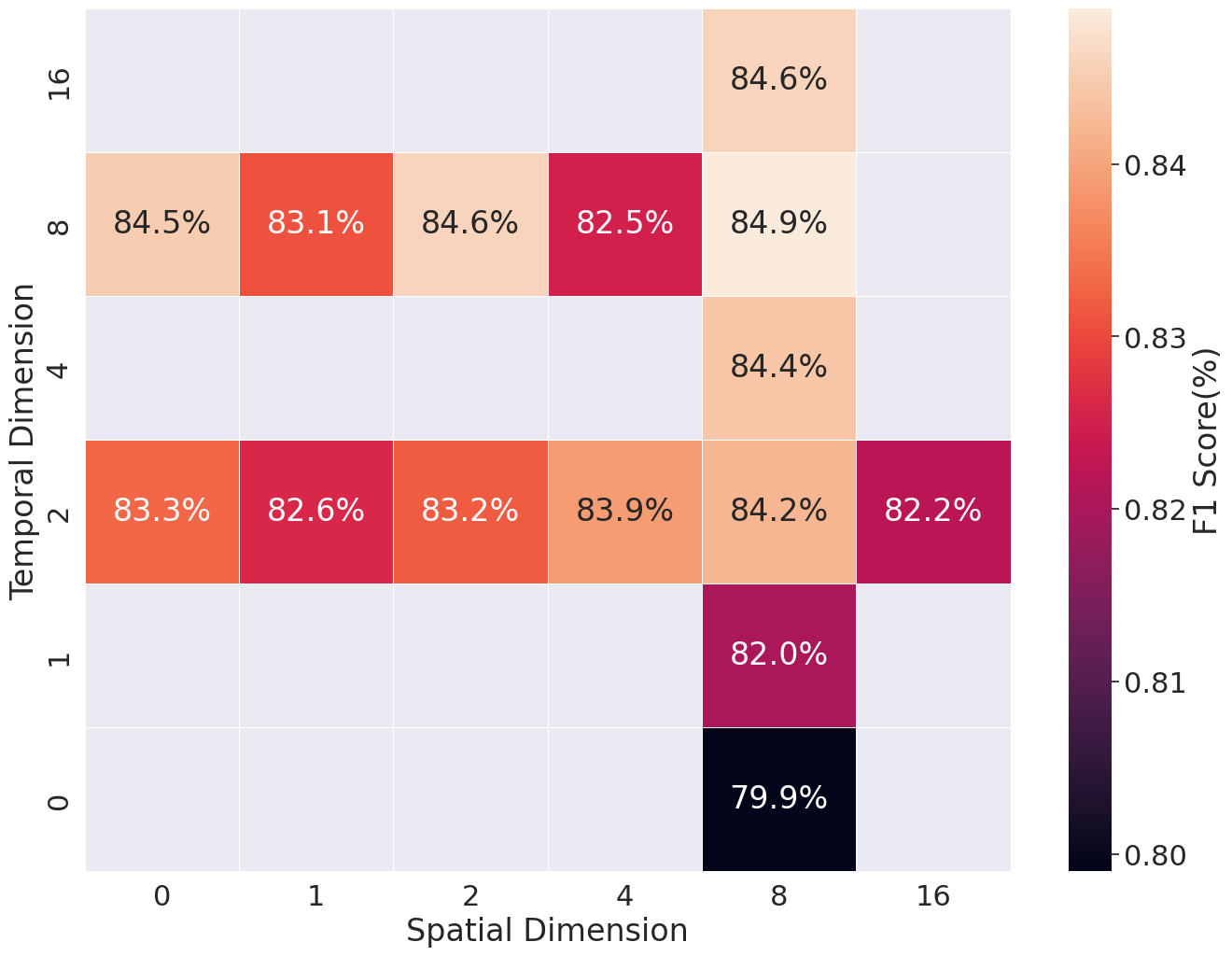}
 \end{center}
    \caption{F1-scores results upon varying AdapterFromer's spatial and temporal dimensions  on Beqaa-Lebanon dataset.}
 \label{adaptformer_results_fig}
 \end{figure}

\textbf{Visual Prompt Tuning} experiments delves further into the impact of various configurations. We conducted four series of experiments to assess the influence of different configuration settings, as shown in Table\ref{table:vpt}.

\begin{itemize}
    \item \textbf{Series 1}: External prompt is not used and the model is deep. The F1 score hovers around the low 82\% range, regardless of the temporal and spatial dimensions.
    \item \textbf{Series 2}: We used an external deep prompt that shows outstanding performance, where the highest F1-score of 83.5\% is achieved at dimensions $16*16$.
    \item \textbf{Series 3}: We used an external shallow prompt and witnessed a dip in performance, with F1 scores of 81.36\%  and 79\% for dimensions $8*8$ and $4*4$, respectively. This suggests that although the shallow prompt might not be as effective as the deep prompt, it still outperforms head tuning, as shown in Table~\ref{table:general_comparison}.
    \item \textbf{Series 4}: We experiment here with setting one dimension (temporal and spatial) at a time to zero while using an external deep prompt. An F1-score of 83.5\% is achieved when the temporal dimension is set to 8 and the spatial dimension is equal to zero, but the performance drops significantly to 69.73\% the other way around (temporal set to zero).
\end{itemize}

\textbf{Series 1 and 2} showed that using the prompt inside the transformer's $Q$ and $K$ matrices only, which was proposed in the literature, does not have any advantage over adding it outside the transformer encoder layer. On the other hand, \textbf{Series 2 and 3} showed that deep prompts lead to better performance. \textbf{Series 4} concludes that prompt tuning of only the temporal transformer is sufficient, and it is safe not to tune the spatial transformer.\\

Similarly to VPT, we test the influence of the \textbf{Adaptformer} spatial and temporal dimensions as shown in Figure \ref{adaptformer_results_fig}. Spatial adapters showed negligible importance compared to temporal ones. A \textbf{84.5\%} F1-score is attained when we set the temporal dimension to 8 and ignore the spatial dimension. On the contrary, we witness a \textbf{4.6\%} drop in F1-score when doing the opposite (setting the temporal dimension to zero). AdaptFormer revealed the best performance among all PEFT techniques and the baselines with a F1-score of \textbf{84.9\%} when the adapter temporal and spatial dimensions were equal to 8.\\

We finally note that LoRA and Adaptformer were the only PEFT techniques that provided better F1-scores (84.76\% and 85\%, respectively) than full fining of the model while training only less than 1\% of the model parameters. Also, as tabulated in Table \ref{table:general_comparison}, both Bitfit and VPT provided comparable but fewer results than full fine-tuning, while still having at least 27\% better F1-score than the simple Head-tuning baseline. This shows that all considered PEFT techniques, when tuned carefully, are capable of reaching 98\% of fully trained model performance. Furthermore, we investigate the effect of varying the learning rate in all PEFT methods. These experiments will not be shown here because of space limitations.

In summary, following an extensive set of experiments, AdaptFormer model produced an 85\% F1-score on the Beqaa-Lebanon test set with only 1.09\% of the model parameters.

\subsection{Munich 480 dataset}
To further support our findings, we apply the following five training techniques in the Munich 480 dataset: \textit{(i)} Head Tune, \textit{(ii)} Partial Token Tune, \textit{(iii)} Token Tune, \textit{(iv)} AdaptFormer and \textit{(v)} Training from scratch as shown in Table \ref{munich480_results}.

Partial and full token tuning provided better F1 scores (67.4\% and 75.2\%, respectively) than head tuning (63.9\%). Adaptformer outperforms partial and full token tuning techniques with a margin greater than 10\%. Moreover, Adaptformer only lays 4\% behind full model training while only training 0.8\%of the model parameters.

\begin{table}[h]
\centering
\begin{tabular}{|c|c|c|c|}
 \hline
 Training Technique & F1-score (\%) & IoU (\%) & Trainable Parameters (\%)\\
 \hline
  AdaptFormer & 84.7 & 74.3 & 0.808\\
  \hline \midrule
 Head Tune & 63.9 & 48.4 & 0.079\\
 \hline
 Partial Token Tune & 67.4 & 52.1 & 0.061\\
 \hline
 Full  Token Tune & 75.2 & 61.3 & 0.208\\
 \hline
 Training from scratch & 88.9 & 80.7 & 100 \\
 \hline
\end{tabular}
\caption{F1-scores of different investigated training scenarios in the Munich 480 dataset.}
\label{munich480_results}
\end{table}

\section{Conclusion}

In this paper, we empirically studied the use of the parameter-efficient fine tuning (PEFT) technique to adapt the state-of-the-art TSViT model in two different countries, namely Lebanon and Germany. We showed that PEFT methods can achieve accurate results with a size of about 1\% of the model parameters. Our investigation provided several important findings on the effectiveness of applying parameter-efficient fine-tuning strategies in TSViT for winter wheat segmentation, where we were able to achieve an F1-score of 85\% while training only 0.72\% parameters. We analyzed each PEFT method's hyperparameters, and pointed out the best settings to achieve the highest possible individual performance.

\vspace{6pt}




\funding{``This research received no external funding''.}

\conflictsofinterest{``The authors declare no conflict of interest.''}

\begin{adjustwidth}{-\extralength}{0cm}

\reftitle{References}


\bibliography{references_TSViT}


\end{adjustwidth}

\end{document}